%% file: main.tex
\documentclass[runningheads]{llncs}

\usepackage[arxiv,year=2026,ID=8234]{eccv}

\usepackage{eccvabbrv}

\usepackage{graphicx}
\usepackage{booktabs}
\usepackage{tabularx}
\usepackage{wrapfig}
\usepackage{xcolor}
\usepackage{pifont}
\newcommand{\cmark}{\scalebox{1.3}{\textcolor{green!70!black}{\ding{51}}}}
\newcommand{\xmark}{\scalebox{1.3}{\textcolor{red!90!black}{\ding{55}}}}

\usepackage{color, colortbl}

\usepackage[accsupp]{axessibility}  

\usepackage{hyperref}

\usepackage{orcidlink}

\newcommand{\Project}{ASAP}

\newcounter{BHNumberOfComments}
\stepcounter{BHNumberOfComments}

\begin{document}

\title{ASAP: Attention-Shift-Aware Pruning for Efficient LVLM Inference} 


\author{Surendra Pathak \and
Bo Han}

\authorrunning{Pathak et al.}

\institute{George Mason University \\
Fairfax, VA, USA \\
\email{\{spathak8, bohan\}@gmu.edu}}

\maketitle

\input{content/00_abstract}

\input{content/01_intro}

\input{content/02_related}

\input{content/03_method}

\input{content/04_results}

\input{content/10_conclusion}

%
%
\bibliographystyle{splncs04}
\bibliography{main}

\appendix
\input{content/11_supplementary}
\end{document}

%% file: content/00_abstract.tex
\begin{abstract}

While Large Vision-Language Models (LVLMs) demonstrate exceptional multi-modal capabilities, the quadratic computational cost of processing high-resolution visual tokens remains a critical bottleneck. Though recent token reduction strategies attempt to accelerate inference, such methods inadequately exploit attention values and fail to address token redundancy. More critically, they overlook the ``attention shift'' phenomenon inherent in LVLMs, which skews token attention scores. In this work, we propose ASAP, a novel training-free, KV-Cache-compatible pruning recipe that comprehensively addresses these limitations. First, we mitigate the attention shift by utilizing a dynamic bidirectional soft attention mask, ensuring the selection of genuinely informative tokens rather than naive attention-based selection. Second, we posit that high semantic redundancy within the token set degrades performance. We therefore introduce a weighted soft merging component that merges semantically similar tokens, preserving only the most feature-dense visual patches for subsequent layers. ASAP achieves virtually lossless compression of visual context, retaining 99.02\% of the original LLaVA-NeXT-7B performance while aggressively slashing computational FLOPs by $\sim$80\%.


\keywords{ Large Vision-Language Models \and Efficient Inference }
\end{abstract}

%% file: content/01_intro.tex
\section{Introduction}
\label{sec:intro}

Large Language Models (LLMs) have emerged as the dominant paradigm for a wide range of natural language processing tasks, demonstrating remarkable capabilities in reasoning, generation, and understanding~\cite{chiang2023vicuna, touvron2023llama, bai2023qwen, touvron2023llama2, jiang2024mistral}. This success has catalyzed a shift toward the development of multimodal systems such as Large Vision Language Models (LVLMs)~\cite{achiam2023gpt, cheng2024videollama, lin2024video, liu2024llavanext, team2024gemini, guo2024llava, bai2025qwen2, zhang2024video}. By integrating visual encoders with LLM backbones, LVLMs have achieved unprecedented performance on challenging vision-and-language tasks, including visual question answering (VQA), image captioning, and multimodal reasoning. However, the adoption of these models is constrained by their enormous computational demands. In particular, the length of visual tokens increases drastically in LVLMs, introducing a computational burden that scales massively with sequence length. For example, while LLaVA-1.5 converts an image into $576$ tokens, LLaVA-NeXT partitions a high-resolution image into four tiles, resulting in a total sequence of $\sim 2,304$ visual tokens. The abundance of visual tokens becomes a bottleneck for Transformer~\cite{vaswani2017attention} during inference, since the computational burden scales quadratically with the sequence length. This challenge has sparked a surge of research toward efficient inference, with token pruning as a promising candidate solution.

\begin{wrapfigure}{r}{0.5\textwidth} 
    \centering
    \vspace{-0.32in}
    \includegraphics[width=\linewidth]{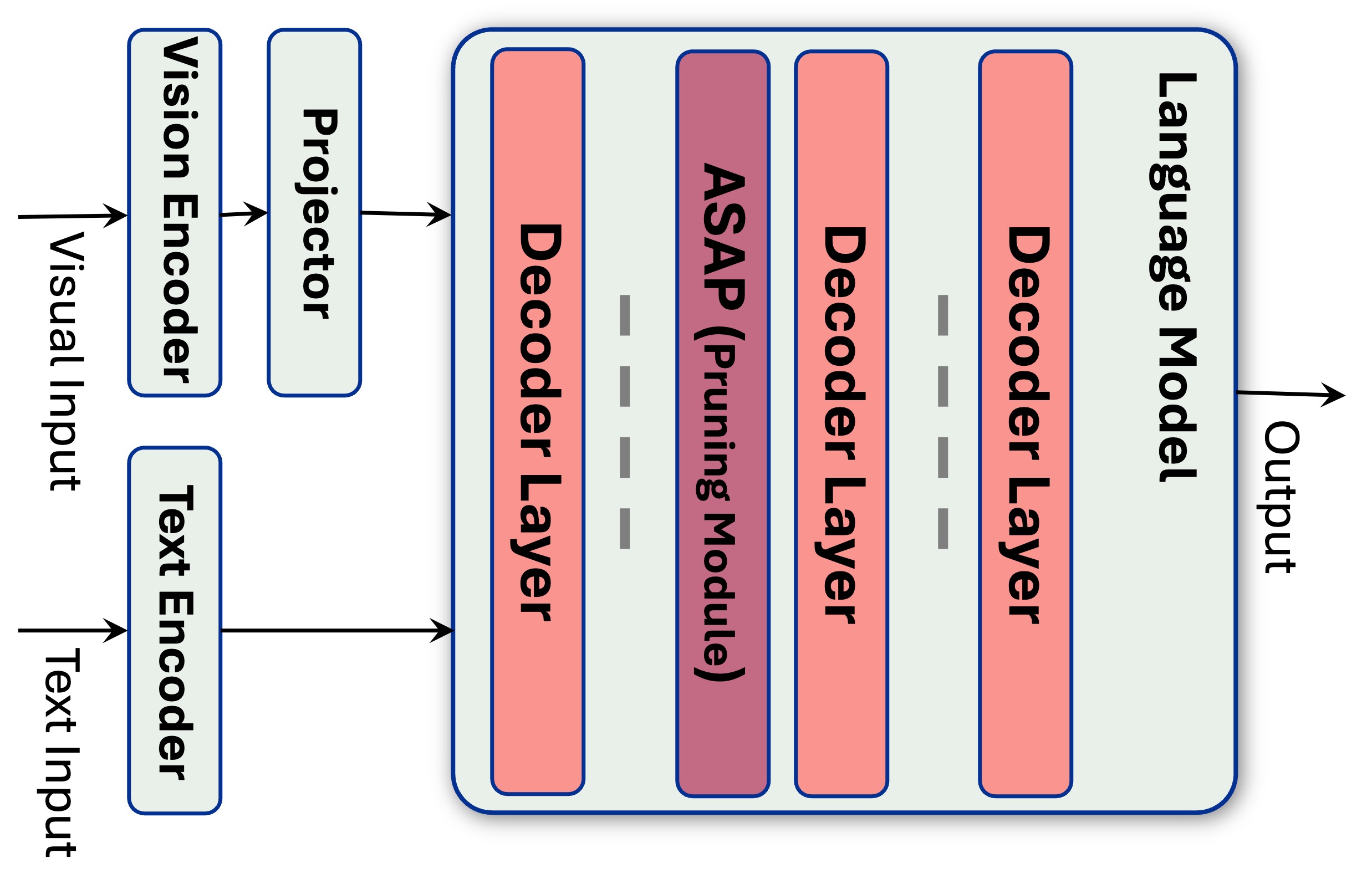}
    \caption{\textbf{ASAP integration within the LVLM architecture.} The plug-and-play pruning module is embedded directly within the language model backbone, operating between standard decoder layers to compress the visual token sequence during the inference forward pass.}
    \vspace{-0.3in}
    \label{fig:premilinary-results}
\end{wrapfigure}

Existing token pruning methods for LVLMs fall into two categories. The first comprises techniques that prune tokens within the vision encoder ~\cite{shang2025llava, yang2025visionzip, zhang2025beyond, arif2025hired}. These methods typically use ViT-attention or feature values to identify candidate tokens for pruning or merging before being processed by the LLM backbone. However, this approach is compromised by a flaw in its text-agnostic nature. Here, pruning decisions are made in a contextual vacuum, lacking information about the user's query for the image, which may lead to degraded model performance and unexpected model responses.

The second category of methods performs pruning within the multimodal LLM backbone~\cite{chen2024image, xing2024pyramiddrop, zhang2024sparsevlm, ye2025fit}. These techniques can access text-vision cross-attention values to perform context-aware, prompt-guided token pruning. Although these techniques can leverage users' prompts, extant works overlook the critical ``attention shift'' phenomenon inherent to LVLMs. Due to the distance decay property inherent to Rotary Position Embeddings (RoPE)~\cite{su2023roformerenhancedtransformerrotary}, later visual tokens receive high attention scores because of architectural artifacts rather than the token's semantic content. Thus, techniques that equate attention values directly with token salience may be affected by this phenomenon, yielding sub-optimal pruning outcomes. While some approaches attempt to circumvent this artifact by completely removing RoPE~\cite{endo2025feather}, such structural modifications strip the model of essential positional priors, degrading its spatial reasoning and accuracy. In this work, we present a fresh perspective to mitigate the attention shift by relaxing the models' standard causal attention. Specifically, we use a soft bidirectional attention mechanism that allows earlier visual tokens to ``peek'' at the later ones,  aggregating context for holistic 2D scene comprehension. In this work, we postulate that an effective pruning strategy must address two distinct issues: the positional bias of attention scores and the inherent redundancy of visual information. Moreover, an image may contain regions of low-semantic-density (e.g., sky, ocean, or forest) that are represented by numerous redundant patch tokens. We argue that an effective pruner must select not only a sparse set of tokens but also ensure that they are feature-dense.

In this work, we propose ASAP, a training-free token pruning framework that maximizes computational savings while maintaining full compatibility with key-value (KV) cache and multi-turn conversations. ASAP consists of three major components. The first component performs attention-shift-aware pruning by leveraging bidirectional attention values, mitigating the influence of the attention shift. The second component eliminates redundant tokens by employing a weighted feature-based similarity metric, thereby preserving a feature-dense token set. Finally, the third component maximizes information retention by salvaging high-salience tokens from the pruned pool to fill the budget slots vacated during the redundancy elimination phase. Unlike prior works, ASAP structurally mitigates attention shift and utilizes a comprehensive three-stage framework for effective token pruning.

Our experimental results demonstrate the superiority of ASAP. When applied to the LLaVA-1.5~\cite{liu2023visual} and LLaVA-NeXT~\cite{liu2024llavanext} models, our method achieves lossless compression, surpassing existing baselines~\cite{chen2024image, shang2025llava, alvar2025divprune, zhang2024sparsevlm}.  Notably, ASAP retains over 99.02\% of the vanilla LLaVA-NeXT-7B model's performance while using only 21.19\% of the original FLOPs, and even achieves supra-vanilla performance on a few benchmarks (e.g., MMBench~\cite{liu2024mmbench}). This work demonstrates that a carefully designed pruning strategy, aware of both shift characteristics and redundancy, can deliver substantial efficiency gains without compromising performance. 

Our contributions are summarized as follows:

\begin{itemize}
    \item We systematically identify and attribute the attention shift artifact in LVLMs to the distance decay property of RoPE. 
    \item To our knowledge, we are the first to structurally mitigate the attention shift artifact, a critical failure point for attention-based token pruning in LVLMs, by employing a soft bidirectional attention mechanism.
    \item We propose ASAP, a novel, training-free pruning framework that co-designs an attention-shift-aware pruning module and a feature-based, weighted redundancy elimination mechanism. Furthermore, ASAP maintains full compatibility with standard KV caching and supports multi-turn conversational inference.
    \item We conduct extensive experiments, demonstrating that ASAP achieves significantly superior performance and outperforms existing baselines. ASAP retains $>$99\% of vanilla accuracy while reducing FLOPs by $\sim$80\%, compared to the standard LLaVA-NeXT-7b model. 
\end{itemize}

%% file: content/02_related.tex
\section{Related Work}
\label{sec:related}

\subsection{Large Vision Language Models}
The remarkable capabilities of LLMs~\cite{chiang2023vicuna,touvron2023llama,bai2023qwen,touvron2023llama2,jiang2024mistral} on diverse tasks such as question-answering, reasoning, coding, etc. has inspired the extension to support multimodal data, such as images and videos, in LVLMs~\cite{liu2023visual, bai2023qwen2, liu2024improved}. These LVLMs usually consist of an LLM backbone and a visual encoder~\cite{radford2021learning, dosovitskiy2020image, sun2023eva, zhai2023sigmoid} that embeds visual tokens before passing them to an alignment module, to align with text embeddings. The capabilities of more recent LVLMs~\cite{achiam2023gpt, cheng2024videollama, lin2024video, liu2024llavanext, team2024gemini, guo2024llava, bai2025qwen2, zhang2024video} have been extended to support high-resolution image and video data, resulting in a large number of visual tokens. For example, LLaVA-NeXT~\cite{liu2024llavanext} converts
a high-resolution image into 4 $\times$ 4 tiles, resulting in a
sequence of $\sim$ 2, 304 visual tokens. Similarly, LLaVA-OneVision~\cite{li2024llava} encodes the image into 384 $\times$ 384 tiles, producing up to 8,748 tokens, and InternVL2~\cite{chen2024expanding} partitions an image into 448 $\times$ 448 tiles, resulting in up to 10,240 tokens. However, the massive visual token count incurs a computational burden that scales quadratically, constraining the adaptation of LVLMs. Thus, efficient inference strategies that save computational resources have become increasingly essential.





\subsection{Token Pruning in LVLM/ViT}
Token compression has emerged as a potential solution to the $O(N^2)$ computational complexity of the self-attention mechanism in LVLMs. Token compression can be achieved by either pruning irrelevant tokens or merging semantically similar tokens. These strategies were first extensively studied in the context of  Vision Transformers (ViTs)~\cite{cao2023pumer, chen2023diffrate, fayyaz2022adaptive, haurum2023tokens, rao2021dynamicvit, bolya2022token, feng2023efficient, kim2024token, wei2023joint} before the era of LVLMs. For example, ToMe~\cite{bolya2022token} introduced a bipartite soft matching algorithm to merge tokens based on feature similarity, while EViT~\cite{liang2022not} utilized attention scores between $[CLS]$ token and visual tokens as a proxy of token importance. In the LVLM domain, these concepts have been adapted but face new challenges. In LVLM, techniques such as LLaVA-Mini~\cite{zhang2025llava}, Llavolta~\cite{chen2024efficient}, and others~\cite{ ye2025atp, zhang2025p} integrate pruning mechanisms but require an additional training phase. These approaches incur an additional computational burden, which undermines the primary goal of efficiency and limits their ``plug-and-play'' applicability. Consequently, training-free, plug-and-play approaches have received more attention recently. These methods can be categorized into several subgroups. The first group that includes LLaVA Prumerge~\cite{shang2025llava}, Visionzip~\cite{yang2025visionzip}, VisPruner~\cite{zhang2025beyond}, and HiRED~\cite{arif2025hired} extends the ViT-era logic, relying on the attention scores between a [CLS] token and the visual tokens. However, these pruning strategies rely on the [CLS] token, which is present only in specific architectures, thereby limiting adaptability to different LVLMs. Another line of training-free work, including FastV~\cite{chen2024image}, PyramidDrop~\cite{xing2024pyramiddrop}, SparseVLM~\cite{zhang2024sparsevlm}, and FitPrune~\cite{ye2025fit}, leverages text-visual cross-attention scores to identify salient tokens. However, these techniques suffer from the critical attention shift phenomenon inherent to LVLMs~\cite{su2024roformer}. Due to positional embeddings like RoPE, later tokens receive high attention scores because of architectural functions rather than the token’s semantic content. Our work structurally addresses this limitation, enabling an efficient pruning recipe that rectifies naive attention-based selection.

%% file: content/03_method.tex
\section{Motivation}
\label{sec:motivation}

\begin{figure}[tb]
  \centering
  \begin{subfigure}{0.44\linewidth}
    \centering
    \includegraphics[width=0.9\linewidth]{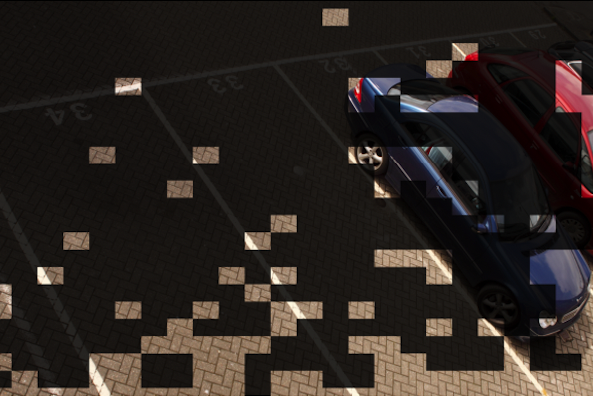}
    \caption{How many cars are there in the parking lot? \\ \textbf{FastV}: There are two cars in the parking lot. \xmark}
    \label{fig:fastv_mask}
  \end{subfigure}
  \hfill
  \begin{subfigure}{0.44\linewidth}
    \centering
    \includegraphics[width=0.9\linewidth]{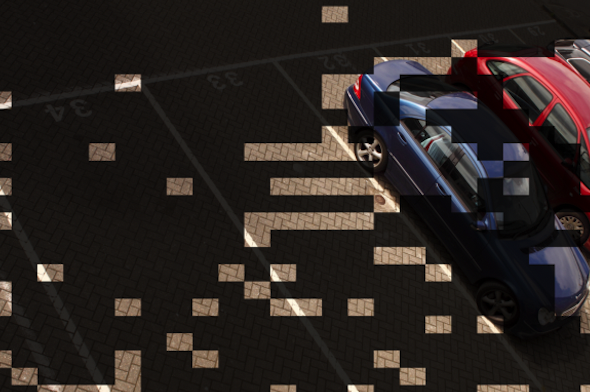}
    \caption{How many cars are there in the parking lot? \\ \textbf{ASAP}: There are three cars in the parking lot. \cmark}
    \label{fig:asap_mask}
  \end{subfigure}
  \caption{\textbf{Qualitative comparison of token selection.} Retained tokens (top 128) are visible, while dropped tokens are darkened. FastV demonstrates a bottom spatial bias, missing the distant third car. ASAP successfully mitigates this bias, preserving critical background features to accurately count all cars.}
    \label{fig:masked_images} 
\end{figure}

To understand the limitations of the current visual token pruning methods, we investigate attention value patterns within the LLM backbones of LVLMs. Our analysis reveals a pronounced spatial bias in which bottom visual tokens consistently receive disproportionately high attention scores across benchmarks, independent of underlying feature salience. Figure~\ref{fig:masked_images} illustrates a representative case: when prompted to count cars, a naive attention-based pruning strategy retains the bottom-most patches, discarding background features and consequently missing the distant third car entirely. This demonstrates that relying on uncalibrated, raw attention scores for token selection is fundamentally flawed. This attention shift is attributed to RoPE, a relative positional encoding mechanism that is used in popular LLMs (e.g., LLaMA~\cite{touvron2023llama}, Qwen~\cite{bai2023qwen}, etc.). RoPE implicitly imposes a distance penalty on tokens positioned further from the text query, inadvertently suppressing early visual features. This suppression underscores the critical need to mitigate positional bias for holistic scene comprehension. In the remainder of this section, we formally attribute this attention shift to the RoPE distance decay mechanism and subsequently propose an architectural approach to mitigate it, restoring balanced visual information routing.

\subsection{Explaining Attention Shift: The RoPE Decay}
We observed the prevalence of attention shift in LVLMs in our previous section. We attribute this bias not to semantic relevance, but to a structural artifact of RoPE. For a query at position $m$ and key at position $n$, RoPE computes their pre-softmax attention score as $s(m, n) = \mathbf{x}_m^\top \mathbf{W}_q^\top \mathbf{R}_{\Theta, m-n} \mathbf{W}_k \mathbf{x}_n$. This formulation inherently imposes a distance decay penalty via the relative rotation matrix $\mathbf{R}_{\Theta, m-n}$, implicitly favoring tokens that are positionally closer to the query. 

In LVLMs, a 2D image is often flattened into a long 1D token sequence during computation. However, this leads to a massive positional gap between the image tokens and the text query. The initial visual tokens (top-left of the image) are positioned significantly further away from the subsequent text queries than the terminal tokens (the bottom-right). Due to this larger distance, RoPE unfairly penalizes the early visual features. The distance decay penalty is suitable for LLMs such as Llama~\cite{touvron2023llama2}, where distant previous text has less importance in subsequent text generation. However, this becomes a limitation in LVLMs, where distant visual features are unfairly penalized.

\subsection{Revisiting Causal Attention for LVLMs }

Beyond the positional bias induced by RoPE, information routing in LVLMs is fundamentally constrained by the standard attention masking strategy. A causal mask $\mathbf{M}_{\text{causal}} \in \mathbb{R}^{N \times N}$ is additively applied to pre-softmax attention logits $\mathbf{S}$  prior to softmax normalization:

\begin{equation}
    \mathbf{A} = \text{Softmax}(\mathbf{S} + \mathbf{M_{\text{causal}}}) .
    \label{eq:causal_attn}
\end{equation}
Because this mask is strictly defined as $M_{\text{causal}_{i,j}} = 0$ for $i \ge j$ and $M_{\text{causal}_{i,j}} = -\infty$ for $i < j$, the attention weight $A_{i,j}$ for any subsequent patch ($j > i$) is driven exactly to zero. By masking future value vectors, this unidirectional constraint ensures the autoregressive property necessary for next-token prediction. This sequential inductive bias is highly effective for natural language tasks, as text inherently unfolds along a temporal axis where historical context predominantly dictates subsequent semantics. However, directly inheriting this masking strategy in LVLMs introduces a critical misalignment, given that images are intrinsically non-sequential, two-dimensional spatial representations. Visual features are structurally interdependent, where the semantics of an early patch frequently depend on the global context of surrounding patches. Consequently, enforcing strict causality artificially fragments these spatial dependencies along an arbitrary 1D serialization axis, depriving early visual tokens of the global receptive field. This necessitates bidirectional information flow for comprehensive scene understanding.


\section{Methodology}
\label{sec:method}
Motivated by the observed structural limitations in Section~\ref{sec:motivation}, we propose a bidirectional attention strategy to allow unrestricted information flow among visual tokens. Specifically, we relax the strict unidirectional constraint of the standard causal attention exclusively for the visual tokens, while preserving standard auto-regressive constraints for all subsequent text tokens. While recent uniform bidirectional masks~\cite{xing2024mitigating, tian2025identifying} relax these constraints, their static nature inadvertently broadcasts noise from uninformative background features. To overcome this, we propose a dynamic, targeted masking strategy to optimize visual information flow.


\subsection{Salience-Guided Bidirectional Attention}
\label{sec_sub:bidirectional_attn}
Motivated by our prior findings in Section~\ref{sec:motivation}, we propose a Salience-Guided Bidirectional Mask (SG-BiMask) that dynamically relaxes the causal constraint for visual tokens to overcome the spatial-temporal discrepancy of strict autoregression. Rather than uniformly blinding queries to future keys, our approach introduces a target-dependent, additive penalty matrix $\mathbf{M}_{\text{bidir}}$ that scales the visibility of a future patch based on its underlying structural importance. We define $\mathbf{W} = \frac{\mathbf{Q}\mathbf{K}^T}{\sqrt{d}} \in \mathbb{R}^{N \times N}$ as the pre-softmax alignment matrix computed during the standard forward pass. To extract the structural salience of each visual token without introducing auxiliary $\mathcal{O}(N^2)$ computation, we repurpose the in-flight pre-softmax alignment matrix $\mathbf{W}$. Let $\mathcal{I}$ denote the set of indices corresponding to the visual patches. We formulate the raw attention mass $s_j$ of a target key token $j \in \mathcal{I}$ as its summed alignment with all visual queries $i \in \mathcal{I}$:

\begin{equation}
    s_j = \sum_{i \in \mathcal{I}} W_{i,j} .
    \label{eq:attn_mass}
\end{equation}

Because pre-softmax logits can operate in arbitrary magnitude spaces depending on the layer depth and model architecture (often resulting in deeply negative absolute values), analyzing them natively obscures their relative importance. To project these raw logits into a bounded salience distribution $\hat{s}_j \in [0, 1]$ while preserving the relative variance learned by the attention mechanism, we apply Min-Max normalization directly to the raw attention mass:

\begin{equation}
    \hat{s}_j = \frac{s_j - s_{\min}}{s_{\max} - s_{\min} + \epsilon}   ,
    \label{eq:salience}
\end{equation}
where $s_{\min}$ and $s_{\max}$ represent the minimum and maximum floored masses across the image span, and $\epsilon = 10^{-6}$ ensures numerical stability.

This normalized salience explicitly dictates the bidirectional leak budget for each token. We define the attention discount factor $\lambda_j = \lambda_{\max} \cdot \hat{s}_j$, where $\lambda_{\max} \in (0, 1]$ is a configurable hyperparameter governing the maximum allowable forward visibility. Because the mask is additively combined with the logits prior to the exponential softmax operation, this scalar must be projected into log-space. The proposed bidirectional mask $\mathbf{M}_{\text{bidir}}$ is formally defined to relax the causal constraint by overwriting the $-\infty$ values in the upper triangle ($i < j$) of the image span with our salience-dependent penalty:

\begin{equation}
    M_{\text{bidir}_{i,j}} = \begin{cases} 0 & \text{if } i \ge j \\ \ln(\max(\lambda_j, \epsilon)) & \text{if } i < j \end{cases} .
    \label{eq:bidirectional_mak}
\end{equation}
By replacing $\mathbf{M}_{\text{causal}}$ with $\mathbf{M}_{\text{bidir}}$, the final attention computation becomes:
\begin{equation}
    \mathbf{A} = \text{Softmax}(\mathbf{W} + \mathbf{M}_{\text{bidir}}).
    \label{eq:bidirectional_attention}
\end{equation}

For any previous token $i$ attending to a future target $j$, the exponent rule resolves the additive log-penalty as a scalar on the unnormalized probability: $\exp(W_{i,j} + \ln(\lambda_j)) = \lambda_j \cdot \exp(W_{i,j})$. Consequently, highly salient structural anchors are permitted to broadcast their features backward through the sequence with minimal attenuation, while non-informative background patches are strongly suppressed. By relaxing 1D temporal constraints for salient visual anchors, this mechanism aligns the sequential priors of LLM generation with the spatial concurrency required for holistic 2D scene comprehension.

\begin{figure*}[t] 
    \centering
    \includegraphics[width=1\linewidth]{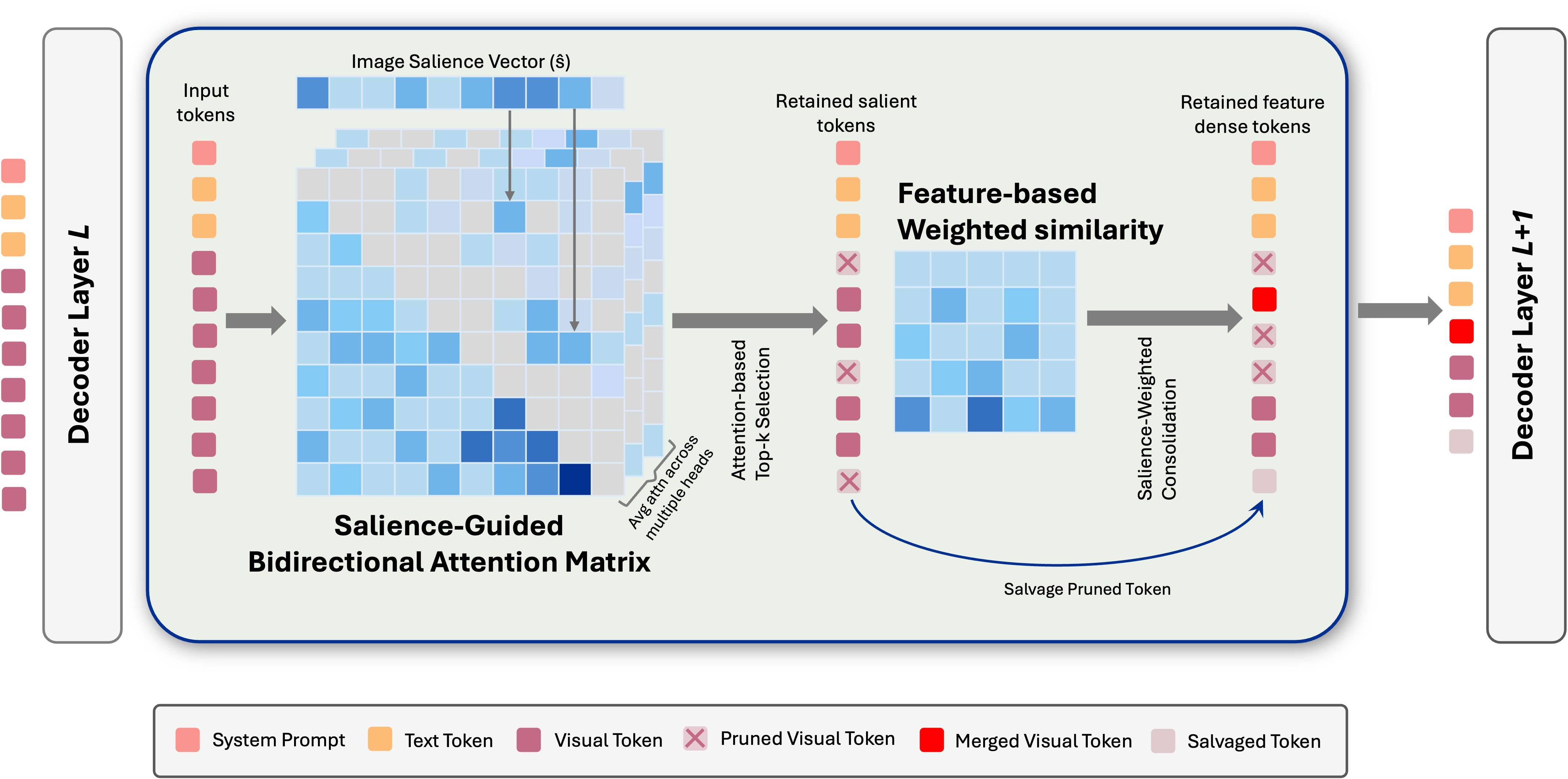}
    \caption{\textbf{The two-stage ASAP architecture.} After decoder layer $L$, a salience-guided bidirectional matrix retains top-$k$ salient visual tokens. A feature-based similarity module then consolidates redundant tokens via salience-weighted merging and a salvage mechanism. This dual-filtering pipeline passes a compact, feature-dense visual set to layer $L+1$, strictly preserving all text and system prompts.}
    \label{fig:detailed-architecture}
\end{figure*}

\subsection{Salience-Weighted Consolidation and Budget Reallocation }
\label{sec_sub:consolidation}

While the \Project{} mask successfully preserves high-impact visual anchors, it inherently leads to the retention of semantically redundant patches (e.g., adjacent patches of a uniform background). Directly applying standard reduction algorithms treats all features uniformly during aggregation~\cite{bolya2022token}. In the context of LVLMs, this cardinality-based averaging risks severe feature dilution, blurring the salient visual patches isolated by our bidirectional mask~\cite{wen-etal-2025-stop}.

To optimize the informational density of our predefined token retention budget (defined in Section~\ref{sec_sub_main_results},~\ref{sec_sub_hd_results}, and~\ref{sec_sub_internvl_results}, respectively), we introduce a Salience-Weighted Consolidation mechanism. This module couples feature redundancy with the structural importance scores ($\hat{s}$) extracted during the Salience-Guided Bidirectional Masking phase in Section~\ref{sec_sub:bidirectional_attn}.

Given the preserved token set $\mathcal{I}_{\text{keep}}$, we identify semantic redundancy by computing the pairwise cosine similarity $S_{i,j}$ between hidden states. Rather than applying a naive average or summation, we aggregate the features using a convex combination weighted by the normalized attention mass $\hat{s}$ derived from Equation~\ref{eq:salience}:

\begin{equation}
    h_i \leftarrow \frac{\hat{s}_i h_i + \sum_{j \in \mathcal{I}_{\text{src}}(i)} \hat{s}_j h_j}{\hat{s}_i + \sum_{j \in \mathcal{I}_{\text{src}}(i)} \hat{s}_j}  .
    \label{eq:weighted_merging}
\end{equation}

To prevent cyclic dependencies during aggregation, we introduce a strict salience-anchored routing: a source token $j$ is absorbed by a destination token $i$ if and only if $S_{i,j} > t$ and $\hat{s}_i \geq \hat{s}_j$ (breaking ties by sequence index). This directional mapping establishes disjoint source sets $\mathcal{I}_{\text{src}}(i)$, guaranteeing that feature consolidation strictly flows toward the most structurally significant anchors.

This formulation ensures that when redundant tokens are fused, the resulting latent representation is heavily anchored by the patch that the previous module deemed most structurally critical, preventing feature dilution.

The fusion of $a := |\bigcup_i \mathcal{I}_{\text{src}}(i)|$ source tokens vacates $a$ positions within the $k$-token budget. To optimize information retention, we reallocate these available slots to the top $a$ tokens from the previously pruned pool, ranked by their attention mass ${s}_j$. Consequently, this integration of salience-driven fusion and candidate recovery optimizes the fixed $k$-token budget, ensuring that the final sequence maximizes semantic diversity rather than merely truncating spatial representation.

%% file: content/04_results.tex
\section{Experiments}
In this section, we present the results of our experiments that demonstrate the effectiveness and superiority of our framework compared to existing baselines. Importantly, we establish the efficacy of ASAP in mitigating the RoPE decay visual attention shift.

\subsection{Experimental Setup}
\subsubsection{Datasets.}
We conducted our experiments on several commonly used benchmarks for different tasks, including visual question answering, object recognition, optical character recognition, etc. Particularly, we accessed ASAP's results on VQAv2~\cite{goyal2017making}, GQA~\cite{hudson2019gqa},  ScienceQA~\cite{lu2022learn}, TextVQA~\cite{singh2019towards}, MME~\cite{Fu2023MMEAC}, MMBecnch~\cite{liu2024mmbench}, MMBench-CN~\cite{liu2024mmbench}, and MMVet~\cite{yu2023mm}. 
VQAv2 assesses foundational, open-ended visual question answering, while GQA evaluates complex compositional reasoning from scene graph-based structures. TextVQA and ScienceQA assess fine-grained, specialized reasoning. TextVQA measures OCR-based reasoning on text within images, while ScienceQA tests complex scientific reasoning on diagrams. MME, MMVet, and MMBench are comprehensive evaluation benchmarks where MME assesses broad perception and reasoning across 14 sub-tasks, MMVet evaluates intricate real-world reasoning challenges across 6 capabilities and 16 tasks, and both MMBench and MMBench-CN measure fine-grained perception and logic. All our experiments follow the default settings and metrics for each benchmark. 

\subsubsection{Implementation details.}
We implemented ASAP on different LVLMs, namely LLaVA-1.5-13B~\cite{liu2023visual}, and LLaVA-NeXT (LLaVA-1.6-7B)~\cite{liu2024llavanext} to demonstrate the effectiveness of our pruning recipe across different models. LLaVA-1.5-13B encodes a standard $336 \times 336$ image using 576 visual tokens, supporting low-resolution images. On the other hand, LLaVA NeXT supports high-resolution images by partitioning an image into $4 \times 4$ tiles, generating up to 2,880 tokens. This $4 \times$ increase in token count over LLaVA-1.5 presents a significant computational bottleneck, making it an ideal testbed for our pruning framework. In all our experiments, we implement ASAP once at the layer $k=2$ of the corresponding backbone LLM.

\subsubsection{Baselines.}
We compare ASAP with existing token pruning and merging techniques to assess the effectiveness of our strategy. For a comprehensive comparison, we selected a diverse set of baselines, including classic and recent methods, techniques based on either attention or feature similarity, and approaches that operate within either the LLM backbone or the vision encoder. For a fair comparison, all the selected methods are plug-and-play techniques that can be integrated into existing LVLMs without additional training. The selected baselines include FastV~\cite{chen2024image}, LLaVA-PruMerge+~\cite{shang2025llava}, SparseVLM~\cite{zhang2024sparsevlm}, DivPrune~\cite{alvar2025divprune}, and VisPruner~\cite{zhang2025beyond}. FastV and SparseVLM are text-aware pruning strategies that prune visual tokens in the LLM backbone. In contrast to attention-based top-$k$ selection in FastV, SparseVLM selects text raters and adaptively prunes visual tokens.  LLaVA-PruMerge+ and VisPruner leverage \texttt{[CLS]} attention to determine salient visual tokens in the image encoder before passing to the LLM backbone. In contrast, DivPrune formulates token pruning as a ``Max-Min Diversity Problem'' (MMDP) to select a subset of tokens that maximizes feature distance.

\subsection{Main Results}
\label{sec_sub_main_results}
We evaluated ASAP against leading training-free baselines on diverse image understanding tasks. Our evaluation was conducted at two distinct computational budgets, corresponding to $\sim$32.99\% and $\sim$21.89\% of the vanilla LLaVA-1.5-7b's total FLOPs. The procedure to obtain the theoretical FLOPs is outlined in Appendix~\ref{adx:efficiency_evaluation}.  We selected these budgets to assess ASAP's efficiency-to-accuracy tradeoff at moderate and aggressive pruning levels, respectively. To normalize comparisons across all benchmarks, we report an average performance metric, where scores are scaled relative to the vanilla model.

ASAP establishes impressive results in training-free pruning, demonstrating a superior trade-off between computational cost and model performance.  As detailed in Table~\ref{tab:results_llava1.5_13b}, our method is highly effective across diverse pruning budgets. At a high-efficiency setting (retaining 128 tokens at a 20.90\% FLOPs ratio), our method retains 97.10\% of the vanilla model's average performance. Under this extreme compression, ASAP sees minimal degradation on benchmarks such as MMBench (66.18 vs. 67.71) and MMBench-CN (62.46 vs. 63.60) while robustly maintaining performance on TextVQA (58.88 vs 61.19). 

At a more moderate 31.88\% FLOPs ratio (retaining 192 tokens), our model retains an impressive 98.68\% of the original average performance, further validating the framework's efficacy. At this budget, ASAP pushes its supra-vanilla achievements even further, outperforming the full LLaVA-1.5-13B model on MME (1546.40 vs. 1529.93), MMBench (68.29 vs. 67.71), and SQA (72.89 vs. 71.6).

Compared to existing methods, ASAP consistently outperforms the strongest prior baseline, DivPrune~\cite{alvar2025divprune}. ASAP beats DivPrune's average performance retention by 1.08\% at the 128-token budget (97.10\% vs. 96.02\%) and by 2.00\% at the 192-token budget (98.68\% vs. 96.68\%), confirming its definitive superiority in optimizing the efficiency-to-accuracy frontier.

\begin{table*}[t]
    \centering
    \caption{\textbf{Performance comparison of ASAP with other pruning strategies on LLaVA-1.5-13B~\cite{liu2023visual}.} Ratio represents the FLOPs ratio (percentage of vanilla FLOPs retained). Avg refers to the average percentage performance across all benchmarks compared to the vanilla model. Best scores are highlighted in \textbf{\textcolor{red}{red}}.}
    \label{tab:results_llava1.5_13b}
    
    \tiny 
    \renewcommand{\arraystretch}{1.15} 
    \setlength{\tabcolsep}{2pt} 
    
    \newcommand{\best}[1]{\textcolor{red}{\textbf{#1}}}
    
    \begin{tabular}{l|cc|ccccccc|c}
        \toprule
        \textbf{Method} & \textbf{FLOPs} & \textbf{Ratio} & \textbf{VQA$^{\scalebox{0.6}{\text{V2}}}$} & \textbf{GQA} & \textbf{SQA} & \textbf{VQA$^{\scalebox{0.6}{\text{Text}}}$} & \textbf{MME} & \textbf{MMB} & \textbf{MMB$^{\scalebox{0.6}{\text{CN}}}$} & \textbf{Avg}  \\
        \midrule
        LLaVA-1.5-13B & 3.817 & 100\% & 80.0 & 63.25 & 71.6 & 61.19 & 1529.93 & 67.71 & 63.60 & 100 \\
        \midrule
        
        \multicolumn{11}{c}{\cellcolor[gray]{0.88} \textbf{Retain 192 tokens \textcolor{ForestGreen}{($\downarrow$ 66.67\%)}}} \\ 
        \midrule
        FastV {\scalebox{0.8}{ [ECCV'24]}} & 1.259 & 32.99 & 75.42 & 58.67 & 70.64 & 57.23 & 1442.31 & 66.43 & 61.05 & 95.37 \\
        PruMerge+ {\scalebox{0.8}{ [ICCV'25]}} & 1.253 & 32.83 & 76.97 & 59.72 & 71.10 & 57.12 & 1451.31 & 66.32 & 62.54  & 95.94 \\
        SparseVLM {\scalebox{0.8}{ [ICML'25]}} & 1.253 & 32.83 & 78.31 & 57.80 & \best{73.08} & 51.93 & 1485.27 & 67.61 & 61.77  & 95.75 \\
        DivPrune {\scalebox{0.8}{ [CVPR'25]}} & 1.253 & 32.83 & 78.42 & 59.12 & \best{73.08} & 58.84 & 1482.01 & 65.95 & 59.02 & 96.68 \\
        \rowcolor{blue!12} ASAP \scalebox{0.8}{(Ours)} & 1.253 & 31.88 & \best{79.46} & \best{62.43} & 72.89 & \best{60.69} & \best{1546.40} & \best{68.29} & \best{63.40}  & \best{98.68} \\
        \midrule
        
        \multicolumn{11}{c}{\cellcolor[gray]{0.88} \textbf{Retain 128 tokens \textcolor{ForestGreen}{($\downarrow$ 77.78\%)}}} \\ 
        \midrule
        FastV {\scalebox{0.8}{ [ECCV'24]}} & 0.836 & 21.898 & 74.34 & 57.19 & 70.32 & 56.66 & 1411.30 & 64.15 & 60.37 & 93.72 \\
        PruMerge+ {\scalebox{0.8}{ [ICCV'25]}} & 0.833 & 21.801 & 75.78 & 57.32 & 71.89 & 56.50  & 1440.38 & 64.63 & 61.51 & 94.91 \\
        SparseVLM {\scalebox{0.8}{ [ICML'25]}} & 0.833 & 21.801 & 76.18 & 57.85 & \best{73.97} & 53.15 & 1470.43 & 66.09 & 61.51 & 95.55  \\
        DivPrune {\scalebox{0.8}{ [CVPR'25]}} & 0.833 & 21.801 & 76.12 & \best{58.82} & 72.78 & 57.95  & 1460.01 & 65.81 & 60.43 & 96.02 \\
        \rowcolor{blue!12} ASAP \scalebox{0.8}{(Ours)} & 0.797 & 20.901 & \best{76.26} & 58.76 & 72.89 & \best{58.88} & \best{1481.80} & \best{66.18} & \best{62.46} & \best{97.10} \\
        \bottomrule
    \end{tabular}
\end{table*}

\subsection{Results on High-Resolution Images}
\label{sec_sub_hd_results}
To validate the generalizability of ASAP, we integrated it into LLaVA-NeXT, a model designed for high-resolution vision tasks. LLaVA-NeXT partitions input images into $4 \times 4$ tiles, resulting in a large and computationally expensive sequence of up to $\sim 2,880$ visual tokens. While this tiling approach enhances performance on fine-grained tasks, it introduces significant token redundancy, making it an ideal testbed for our method. We evaluated ASAP on LLaVA-NeXT using two distinct retention budgets of  33.33\% and 22.23\%, as previously. In addition to FastV, we use Vispruner~\cite{zhang2025beyond} as a baseline over here since its LLaVA NeXT implementation is available publicly.  As shown in Table~\ref{tab:results_llava_next}, at the 33.33\% budget ($\sim$960 tokens), our method achieves performance that is comparable to, and in some cases surpasses, the vanilla LLaVA-NeXT model. Moreover, with a more aggressive pruning rate retaining only 22.23\% ($\sim$640 tokens) of the original tokens, ASAP's performance degrades by only $\sim$0.98 points on average, demonstrating robust high-compression performance. Furthermore, ASAP establishes a new state-of-the-art, outperforming the VisPruner~\cite{zhang2025beyond} baseline by 1.80\% and 2.32\% at the 33.33\% and 22.23\% retention budgets, respectively.

\begin{table}[t] 
    \centering
    \caption{\textbf{Performance comparison of ASAP with other pruning strategies on LLaVA-NeXT-7B~\cite{liu2024llavanext}.} Avg refers to the average percentage performance across all benchmarks compared to the vanilla model. Best scores are highlighted in \textbf{\textcolor{red}{red}}.}
    \label{tab:results_llava_next}
    
    \scriptsize 
    \renewcommand{\arraystretch}{1.10} 
    \setlength{\tabcolsep}{3.5pt} 
    
    \newcommand{\best}[1]{\textcolor{red}{\textbf{#1}}}
    
    \begin{tabular}{l|c|ccccc|c}
        \toprule
        \textbf{Method} & \textbf{FLOPs} & \textbf{VQA$^{\scalebox{0.6}{\text{V2}}}$} & \textbf{GQA} & \textbf{VQA$^{\scalebox{0.6}{\text{Text}}}$} & \textbf{MMB} & \textbf{MMVet} & \textbf{Avg} \\
        \midrule
        LLaVA-NeXT-7B & 20.825 & 81.2 & 62.9 & 59.6 & 65.8 & 40.0 & 100.0 \\
        \midrule
        
        \multicolumn{8}{c}{\cellcolor[gray]{0.88} \textbf{Retain 960 tokens \textcolor{ForestGreen}{($\downarrow$ 66.67\%)}}} \\ 
        \midrule
        FastV {\scalebox{0.8}{ [ECCV'24]}} & 6.477 & 80.33 & 61.98 & 59.32 & 65.12 & 39.4 & 98.89 \\
        VisPruner {\scalebox{0.8}{ [ICCV'25]}} & 6.458 & 80.0 & 62.20 & 58.81 & 65.29 & 38.4 & 98.26 \\
        \rowcolor{blue!12} ASAP \scalebox{0.8}{(Ours)} & 6.258 & \best{81.19} & \best{62.29} & \best{59.67} & \best{66.41} & \best{40.1} & \best{100.06} \\
        \midrule
        
        \multicolumn{8}{c}{\cellcolor[gray]{0.88} \textbf{Retain 640 tokens \textcolor{ForestGreen}{($\downarrow$ 77.78\%)}}} \\ 
        \midrule
        FastV {\scalebox{0.8}{ [ECCV'24]}} & 4.260 & 77.81 & 61.04 & 58.08 & 64.37 & 37.6 & 96.43 \\
        VisPruner {\scalebox{0.8}{ [ICCV'25]}} & 4.252 & 79.82 & \best{61.22} & 58.34 & 65.29 & 36.3 & 96.70 \\
        \rowcolor{blue!12} ASAP \scalebox{0.8}{(Ours)} & 4.144 & \best{79.91} & 61.17 & \best{58.98} & \best{65.35} & \best{40.48} & \best{99.02} \\
        \bottomrule
    \end{tabular}
\end{table}


\subsection{Results on Different LVLM Architecture}
\label{sec_sub_internvl_results}
To further validate the generalizability of ASAP, we integrated it into InternVL 2.5~\cite{chen2024expanding}, a modern model architecture designed for high-resolution vision tasks. Similar to other high-resolution architectures, InternVL employs a dynamic tiling strategy that generates a massive sequence of visual tokens. While this enhances performance on fine-grained tasks, it also introduces significant token redundancy, making it an ideal testbed for our method. Since other baselines's InternVL 2.5 implementation is not available publicly, we use FastV as the baseline over here. As shown in Table \ref{tab:results_internvl}, with the FLOPS ratio of 31.56\% ($\sim$ 510 tokens), ASAP maintains the model's performance, experiencing a minimal average degradation of only 1.24 points compared to the vanilla model while drastically reducing the computational cost from 12.575 to 3.969 FLOPs. Moreover, ASAP remains robust under a more aggressive pruning rate of 21.17\% FLOPs ratio ($\sim$ 340 tokens) by minimizing the accuracy drop. Furthermore, ASAP establishes superior efficiency over existing approaches, outperforming the FastV~\cite{chen2024image} baseline across average benchmark scores by 1.59\% and 0.67\% at the 31.56\% and 21.17\% retention budgets, respectively. These performances establish the generalizability of the ASAP pruning framework to modern LVLM architectures.

Additionally, ASAP's SG-BiMask component is orthogonal to the existing attention-based pruning methods and can be integrated to increase their pruning effectiveness. The increase in accuracy after integrating SG-BiMask to attention-based pruning is demonstrated in Appendix~\ref{adx:sg-bimask}.

\begin{table}[t] 
    \centering
    \caption{\textbf{Performance comparison of ASAP with other pruning strategies on InternVL 2.5 8B.} Avg refers to the average percentage performance across all benchmarks compared to the vanilla model. Best scores within each token budget are highlighted in \textbf{\textcolor{red}{red}}.}
    \label{tab:results_internvl}
    
    \scriptsize 
    \renewcommand{\arraystretch}{1.10} 
    \setlength{\tabcolsep}{3.0pt} 
    
    \newcommand{\best}[1]{\textcolor{red}{\textbf{#1}}}
    
    \begin{tabular}{l|cc|ccccc|c}
        \toprule
        \textbf{Method} & \textbf{FLOPs} & \textbf{Ratio\scalebox{0.7}{\text{(\%)}}} & \textbf{GQA} & \textbf{SQA$^{\scalebox{0.6}{\text{IMG}}}$} & \textbf{POPE} & \textbf{MME} & \textbf{MMB$^{\scalebox{0.6}{\text{CN}}}$} & \textbf{Avg} \\
        \midrule
        InternVL 2.5 8B & 12.575 & 100.0 & 63.40 & 97.81 & 90.59 & 1696.26 & 82.60 & 100.00 \\
        \midrule
        
        \multicolumn{9}{c}{\cellcolor[gray]{0.88} \textbf{Retain 510 tokens}} \\ 
        \midrule
        FastV {\scalebox{0.8}{ [ECCV'24]}} & 3.969 & 31.56 & 61.44 & 96.03 & 89.10 & 1614.42 & 80.32 & 97.17 \\
        \rowcolor{blue!12} ASAP {\scalebox{0.8}{ (Ours)}} & 3.969 & 31.56 & \best{62.31} & \best{97.07} & \best{90.08} & \best{1645.32} & \best{82.47} & \best{98.76} \\
        \midrule
        
        \multicolumn{9}{c}{\cellcolor[gray]{0.88} \textbf{Retain 340 tokens}} \\ 
        \midrule
        FastV {\scalebox{0.8}{ [ECCV'24]}} & 2.663 & 21.17 & 59.17 & 91.51 & 88.47 & \best{1603.51} & 79.33 & 95.02 \\
        \rowcolor{blue!12} ASAP {\scalebox{0.8}{ (Ours)}} & 2.663 & 21.17 & \best{60.00} & \best{92.36} & \best{89.00} & 1596.91 & \best{80.15} & \best{95.69} \\
        \bottomrule
    \end{tabular}
\end{table}

\subsection{Ablation Study}

\subsubsection{Ablation of framework modules.}
In this section, we conduct a systematic ablation study to decompose the ASAP framework and quantify the contributions of its core modules. All experiments are performed on the LLaVA-1.5-7b model, with a pruning budget of $\sim$128 tokens. The performance is evaluated on the TextVQA and MME benchmarks, as they effectively measure fine-grained reasoning and multimodal comprehension. 

\begin{wrapfigure}{r}{0.5\textwidth} 
    \centering
    \vspace{-0.3in}
    \includegraphics[width=\linewidth]{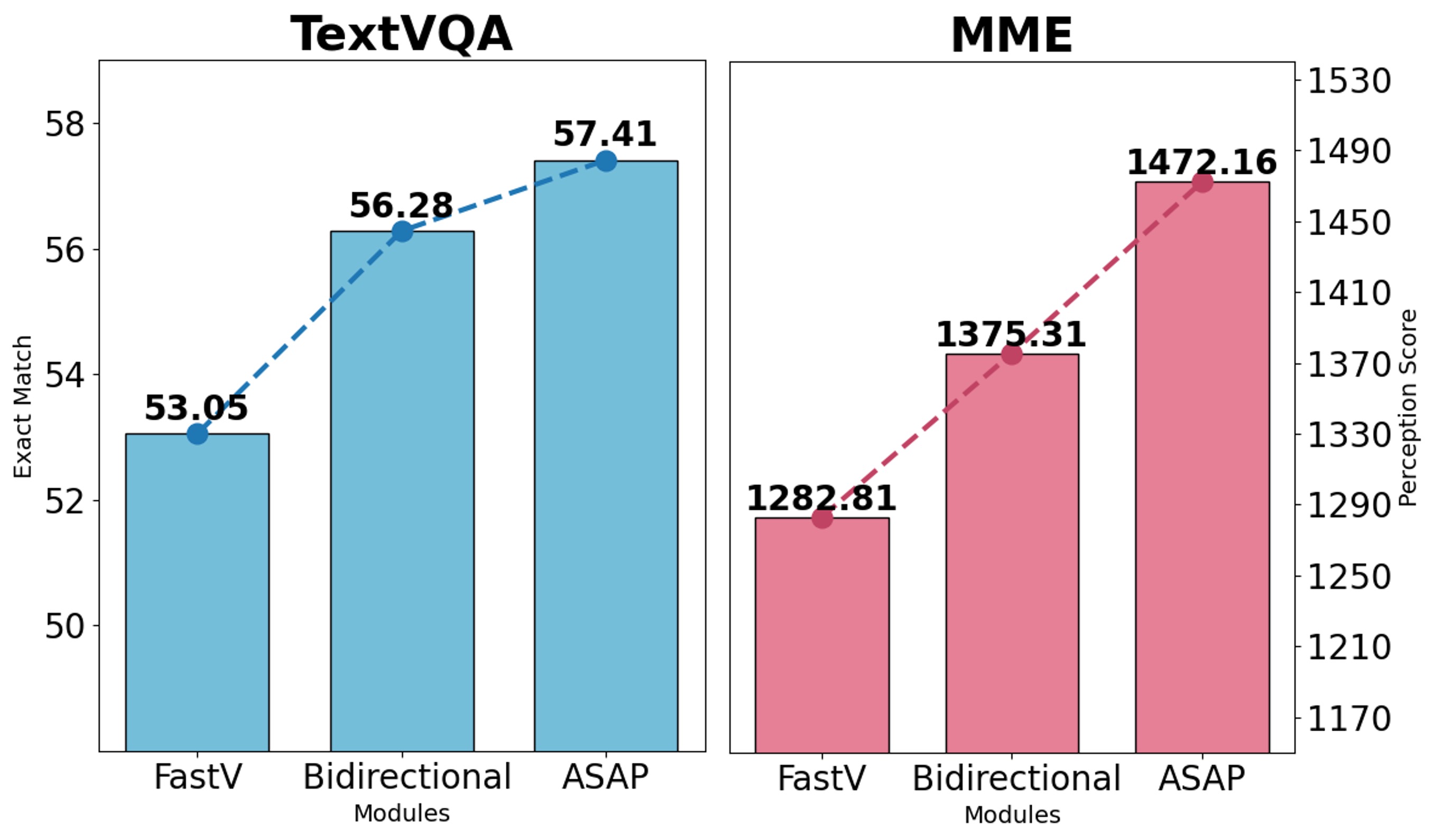}
    \caption{\textbf{ Ablation of ASAP modules on TextVQA and MME.} FastV prunes naively using Top-$k$ selection. Bidirectional uses Salience-Guided Bidirectional attention values. ASAP adds Salience-Weighted Consolidation and Budget Reallocation. The progressive gains confirm that combining attention-driven selection with feature consolidation maximizes semantic density.}
    \vspace{-0.25in}
    \label{fig:modules-ablation}
\end{wrapfigure}

The ablation is structured as an incremental analysis, and the results are summarized in Figure~\ref{fig:modules-ablation}.

\textbf{Baseline (FastV)}: As a starting point, we evaluate the baseline FastV method, which prunes tokens naively using standard causal Top-$k$ selection. Because it lacks bidirectional spatial awareness, this approach achieves base scores of 53.05 on TextVQA and 1282.81 on MME. 

\textbf{Bidirectional (Module 1)}: Next, we replace the standard causal attention with our Salience-Guided Bidirectional attention values for token selection. This provides a significant performance leap, yielding an absolute increase of +3.23 on TextVQA (to 56.28) and +92.50 on MME (to 1375.31) over FastV. This validates the hypothesis that relaxing strict causal constraints and permitting bidirectional information flow allow the model to better identify and retain structurally critical visual anchors.

\textbf{ASAP (Full Framework)}: Finally, we integrate the Salience-Weighted Consolidation and Budget Reallocation modules to form the complete ASAP architecture. Instead of merely discarding unselected tokens, this stage merges semantically redundant features using salience weights and salvages top-ranked pruned tokens to fill the newly freed budget slots. This optimization of the token budget yields an additional performance increase of +1.13 on TextVQA (reaching 57.41) and +96.85 on MME (reaching 1472.16). 

Overall, the progressive gains observed across these stages confirm that combining bidirectional attention-driven selection with salience-weighted feature consolidation successfully maximizes semantic density under a strict token budget.


\subsection{Efficiency Analysis}
\label{adx:interndfsvl}
To evaluate the system efficiency of our approach, we perform a comparative analysis of FLOPs, KV cache storage, GPU memory, and CUDA latency alongside task accuracy against FastV~\cite{chen2024image} on the LLaVA-1.5-7B~\cite{liu2023visual} architecture. As shown in Table \ref{tab:efficiency_llava_1_5}, reducing the visual token count to 128 yields a substantial decrease in computational cost, dropping the required operations from 3.817 to 0.836 TFLOPs. Under equivalent token retention budgets, our method consistently achieves higher accuracy than FastV while simultaneously reducing the KV cache storage footprint. For instance, at the 128-token budget, our approach improves the aggregate accuracy score by over 52 points while requiring approximately 15\% less storage space than FastV. Although our dynamic token selection mechanism introduces a marginal inference overhead of roughly two milliseconds, it provides a strictly superior balance between memory efficiency and downstream model performance.

\begin{table}[t] 
    \centering
    \caption{\textbf{System efficiency comparison on LLaVA-1.5-7B.} We evaluate the computational cost and memory footprint during the prefill phase.}
    \label{tab:efficiency_llava_1_5}
    
    \scriptsize 
    \renewcommand{\arraystretch}{1.0} 
    \setlength{\tabcolsep}{3.0pt} 
    
    \begin{tabular}{l|c|c|c|c|c|c}
        \toprule
        \textbf{Method} & \textbf{\# Token} & \textbf{\shortstack{FLOPs \\ (T)}} & \textbf{\shortstack{Percption \\score}} & \textbf{\shortstack{Storage \\ (MB)}} & \textbf{\shortstack{GPU Memory \\ (GB)}} & \textbf{\shortstack{CUDA Time \\ (ms)}} \\
        \midrule
        LLaVA-1.5-7B & 576 & 3.817 & 1506.46 & 321.39 & 14.33 & 38.27 \\
        \midrule
        
        \multicolumn{7}{c}{\cellcolor[gray]{0.88} \textbf{Retain 192 tokens}} \\ 
        \midrule
        FastV {\scalebox{0.8}{ [ECCV'24]}} & 192 & 1.259 & 1437.45 & 129.73 & 13.98 & 33.65 \\
        \rowcolor{blue!12} ASAP {\scalebox{0.8}{ (Ours)}} & 192 & 1.259 & 1466.19 & 116.64 & 13.96 & 35.13 \\
        \midrule
        
        \multicolumn{7}{c}{\cellcolor[gray]{0.88} \textbf{Retain 128 tokens}} \\ 
        \midrule
        FastV {\scalebox{0.8}{ [ECCV'24]}} & 128 & 0.836 & 1365.435 & 97.33 & 13.97 & 31.45 \\
        \rowcolor{blue!12} ASAP {\scalebox{0.8}{ (Ours)}} & 128 & 0.836 & 1417.524 & 82.14 & 13.96 & 33.45 \\
        \bottomrule
    \end{tabular}
\end{table}

%% file: content/10_conclusion.tex
\section{Conclusion}
\label{sec:conclusion}
In this work, we introduced ASAP, a novel, training-free token pruning framework that establishes a new state-of-the-art efficiency-to-accuracy trade-off for LVLMs. By systematically mitigating RoPE-induced spatial biases and consolidating redundant visual features, ASAP overcomes the critical bottlenecks of existing causal pruning methods. Through extensive experimentation across a wide array of challenging benchmarks, we demonstrated that our approach consistently outperforms established baselines. Notably, ASAP achieves virtually lossless compression—retaining $\sim$99.02\% of LLaVA-NeXT-7B's performance while reducing computational FLOPs by $\sim$80\%.


%% file: content/11_supplementary.tex
\section{Appendix}
\section{\Project{} Efficiency Evaluation} 
\label{adx:efficiency_evaluation}
Presenting only the average number of retained tokens to report pruning efficiency may be insufficient, as it fails to account for the heterogeneous nature of modern pruning strategies. To provide a more standardized and accurate measure of computational cost, we adopt theoretical FLOPs as our primary efficiency metric, a practice consistent with prior work~\cite{chen2024image, xing2024pyramiddrop}. For our evaluation, we calculate the FLOPs attributable to the most computationally intensive components of the LLM backbone: the multi-head attention (MHA) and feed-forward network (FFN) modules. Additionally, we only consider the computation required for the image tokens. The theoretical FLOPs at an LLM layer $L$ is determined as $4vd^2 + 2v^2d + 3vdm$, where $v$ is the total number of image tokens in the layer $L$, $d$ is the hidden state size, and $m$ is the intermediate size of FFN. If the tokens are pruned at multiple layers repeatedly, the total FLOPs is obtained by:


\begin{equation}
    \begin{aligned}
        \Omega(\hat v) &= \sum_{n=0}^{S-1}  L_n \times \left( 4v_n d^2 + 2v_n^2 d + 3v_n d m \right)  \\
        \text{s.t.} \quad v_n &= \left\lceil \lambda_n v\right\rceil, \quad n = 0, 1, \ldots, S-1
    \end{aligned}
    \label{eq:complexity}
\end{equation}
where tokens are pruned $S$ times with respective pruning ratio $\lambda_n$, and $L_n$ is the number of LLM decoder layers with $v_n$ token. The following notation obtains the FLOPs ratio:

\begin{equation}
    \eta = \frac{\Omega(\hat v)}{   L \times\left( 4v d^2 + 2v^2 d + 3v d m \right) } ,
    \label{eq:flops_ratio}
\end{equation}
where $L \times\left( 4v d^2 + 2v^2 d + 3v d m \right)$ represents the FLOPs without applied pruning.

\section{Additional Experiments}
In this section, we present additional experimental results of our ASAP pruning framework to provide further insights on the framework's performance. 

\begin{figure*}[t] 
    \centering
    \includegraphics[width=0.6\linewidth]{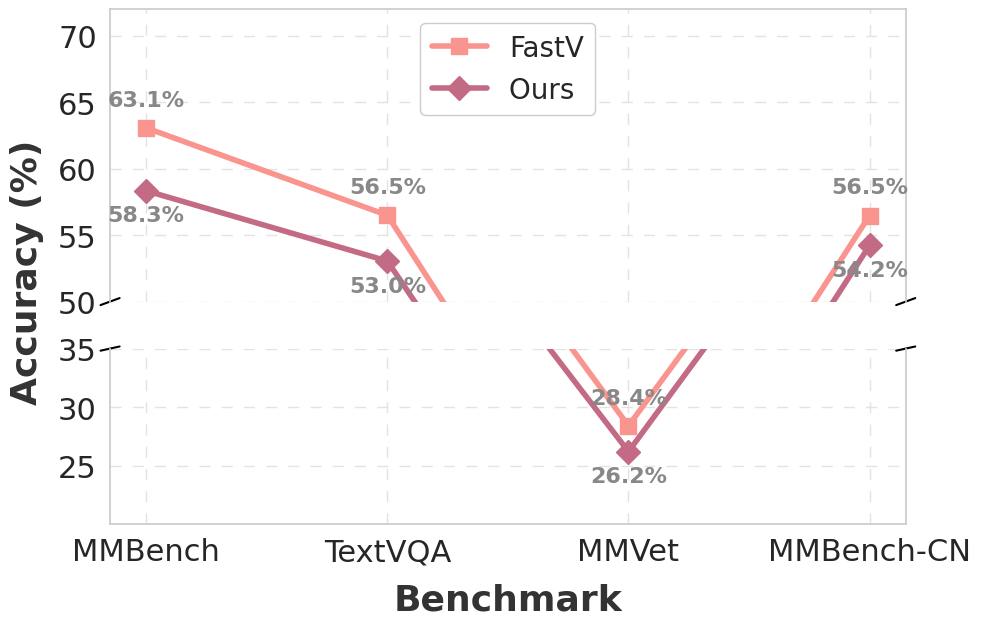}
    \caption{\textbf{Impact of integrating SG-BiMask in existing pruning frameworks.} Our SG-BiMask component consistently improves performance when integrated with the standard causal attention baseline (FastV) across all evaluated benchmarks.}
    \label{fig:premilinary-results}
\end{figure*}

\subsection{SG-BiMask as a Plug-and-Play Component}
\label{adx:sg-bimask}
To demonstrate its modularity, we designed SG-BiMask as a lightweight, plug-and-play component capable of seamlessly upgrading existing LVLM inference pipelines. We empirically validate this versatility by directly integrating it into FastV~\cite{chen2024image}, a widely adopted baseline that conventionally relies on strict causal attention for top-$k$ token pruning. For this experiment, we used the LLaVA-1.5-7B model with the pruning budget of $\sim$128 tokens (FLOPs: 0.797). As Figure~\ref{fig:premilinary-results} illustrates, augmenting FastV with our bidirectional mask immediately yields consistent performance gains across multiple benchmarks. By allowing visual tokens to bidirectionally attend to one another regardless of 1D sequence position, this drop-in module instantly cultivates a richer, spatially aware representation prior to compression.